\documentclass{article}

\PassOptionsToPackage{numbers, compress}{natbib}
\usepackage[preprint]{neurips_2026}

\usepackage[utf8]{inputenc} 
\usepackage[T1]{fontenc}    
\usepackage{hyperref}       
\usepackage{url}            
\usepackage{booktabs}       
\usepackage{amsfonts}       
\usepackage{nicefrac}       
\usepackage{microtype}      
\usepackage{xcolor}         
\usepackage{colortbl}       

\usepackage{soul}
\usepackage{tcolorbox}
\usepackage{color}
\usepackage{marvosym}
\tcbuselibrary{breakable}
\usepackage{multirow}
\usepackage{threeparttable}
\usepackage{amssymb}
\usepackage{pifont}
\usepackage{couriers}
\usepackage{subcaption}
\usepackage{amsmath}
\usepackage{wrapfig}
\usepackage{graphicx}

\title{PercepCap: Video Captioner with Structured Spatio-Temporal Perception}

\author{
  \bfseries Yifan Xu$^{1*\ddagger}$,
  Zihao Wang$^{2\ddagger}$,
  Zhixiao Wang$^{1\ddagger}$,
  Jiaming Zhang$^{1*\ddagger}$,
  Yichun Yang$^1$,
  \\ \bfseries Desen Meng$^1$,
  Yuanxing Zhang$^2$,
  Pengfei Wan$^2$,
  Limin Wang$^{1,3\dagger}$
  \\[0.12cm] $^1$State Key Laboratory for Novel Software Technology, Nanjing Univerisity
  \\ $^2$Kling Team, Kuaishou Technology\quad$^3$Shanghai AI Laboratory
}

\begin{document}

\maketitle

\begingroup
\renewcommand{\thefootnote}{}
\makeatletter\def\Hy@Warning#1{}\makeatother
\footnotetext{$^*$ This work was conducted during the authors' internships at Kling Team, Kuaishou Technology}
\footnotetext{$^\ddagger$ Equal contribution}
\footnotetext{$\dagger$ Corresponding author: Limin Wang}
\endgroup

\begin{abstract}
Video captioning requires fine-grained spatio-temporal understanding of videos, including spatial perception of where objects are located and temporal perception of when events occur.
Existing Multi-modal Large Language Models (MLLMs) usually generate captions directly from video inputs without exposing the perceptual evidence behind their descriptions. As a result, mistakes in spatiotemporal perception are only observed in the final caption, making it difficult to identify and supervise the underlying perceptual errors directly.
To address these issues, we present \textbf{PercepCap}, a perception-aware video captioning framework that makes perceptual evidence explicit before producing the final caption.
Specifically, PercepCap follows a \emph{perceive--describe} generation chain, where the model first produces a spatio-temporal perception trace comprising object trajectories and temporal events according to the video, and then generates the final caption conditioned on the perceived evidence.
To support this new generation paradigm, we design a two-stage training strategy for PercepCap.
\textbf{Perceive-then-Describe Supervised Fine-tuning (PD-SFT)} adapts the model from caption-only generation to the proposed \emph{perceive--describe} chain, while \textbf{Perception-Grounded Reinforcement Learning (PG-RL)} further optimizes perception trace and caption quality with joint rewards over object tracking, temporal event, and object/action description coverage. To support our two-stage training, we introduce
\textbf{Caption-Anchored Perception Data Construction}. This pipeline builds the SFT and RL training data by first generating a caption-only description, extracting the objects and events it mentions, and grounding them back in the video with boxes and timestamps.
This yields caption-aligned perception data that provides solid training ground truth, ensuring that the explicit perception trace and final caption refer to the same objects and events.
Across direct caption evaluation such as DREAM-1K, CaReBench and VidCapBench, and caption-to-QA evaluation like ShortVidBench and MotionBench, PercepCap consistently improves upon the Qwen3-VL baseline and demonstrates leading caption quality.
\end{abstract}

\section{Introduction}
\label{sec:intro}

Detailed video captioning~\cite{dream1k,vidcapbench,carebench,timechatcaptioner,owlcap,vidbridger1} aims to generate fine-grained, temporally organized descriptions that faithfully capture the visual content and evolving events of a video. Achieving this goal requires a model to \emph{perceive} the spatial locations of objects, track the temporal progression of events as they unfold, and translate such evidence into coherent language without losing object--event relations. Recent Multimodal Large Language Models (MLLMs)~\cite{qwen3technicalreport,qwen3.5,Qwen2.5-VL,wang2024qwen2vl,kimivl,glm45v} have substantially advanced open-ended video understanding, yet their descriptions may still omit important objects and events or mischaracterize them.

Despite the importance of perceptual evidence, most existing MLLM-based captioners follow a \emph{caption-only generation paradigm}: they map video inputs directly to final captions, requiring the model to identify relevant objects, actions, and temporal relations and verbalize them within the same autoregressive generation process. This design leaves the perceptual stage largely implicit, giving the model little opportunity to explicitly examine key evidence before producing the caption. Consequently, errors in object identity, action recognition, or temporal ordering appear only in the final text, without revealing which perceptual evidence was missed or misinterpreted. Moreover, when supervision are assigned only to the final caption, the model receives no direct feedback on whether its underlying perception is correct, nor on how perceptual errors should be corrected.

These limitations suggest that detailed captioning should expose video perception evidence inside the generation process. A useful precedent comes from video question answering: VideoP2R~\cite{videop2r} treats video reasoning as a process that first produces an observation of relevant visual evidence and then reasons from that observation to answer questions. This separation makes the intermediate evidence more explicit and allows perception and reasoning to receive more targeted training signals. Inspired by this design, we propose \textbf{PercepCap}, a perception-aware video captioning framework tailored to a different goal: generating a structured detailed perception chain rather than a rough observation. PercepCap converts the intermediate evidence into a \textbf{spatio-temporal, structured perception trace} for captioning. Specifically, PercepCap follows a \emph{perceive--describe} chain: the model first generates a structured perception trace that identifies objects and follows their trajectories over time, while also describing events and localizing their temporal spans. It then generates the final caption conditioned on this perceived evidence. This design separates explicit object- and event-level perception from final caption generation, allowing object trajectories, event boundaries, and final captions to receive distinct learning signals.

Instantiating the \emph{perceive--describe} chain raises two practical challenges. (i)~\textbf{A single supervision signal for different subtasks.} Although supervised fine-tuning can teach the model to generate a perception trace before the caption, it still trains the model to imitate the entire serialized output token by token. In this process, object tracking, event localization, and caption generation are all supervised in the same way: by matching the next token in the reference sequence. The model is therefore not explicitly told whether an error comes from missing an object, assigning a wrong temporal span to an event, or verbalizing the evidence poorly. This makes it difficult to give separate feedback to the perception trace and the final caption, motivating the need for component-specific rewards during reinforcement learning. (ii)~\textbf{Lack of caption-aligned perception annotations.} Standard caption datasets provide final captions but rarely annotate the intermediate evidence that supports them, such as which objects are mentioned, where they appear over time, or when the described events start and end. This makes it difficult to train a structured perception trace that is aligned with the same entities and events expressed in the final caption.

We instantiate PercepCap with a two-stage reinforcement fine-tuning (RFT)~\cite{DBLP:conf/acl/TrungZJSJL24} pipeline that supplies caption-aligned perception data and optimizes it to follow the proposed generation chain. \textbf{Perceive-then-Describe Supervised Fine-tuning (PD-SFT)} adapts the model from caption-only generation to the \emph{perceive--describe} chain, teaching it to produce perception traces before the final caption. \textbf{Perception-Grounded Reinforcement Learning (PG-RL)} then improves the model within this chain using Group Relative Policy Optimization (GRPO)~\cite{DBLP:journals/corr/abs-2501-12948}. In this stage, we use serveral rewards that scores spatial tracking and temporal event grounding in perception chain and object/event coverage in final captions. To support both stages, we introduce \textbf{Caption-Anchored Perception Data Construction}, which first obtains a caption, extracts its object and event content, and then revisits the video to derive object bounding boxes and event timestamps. The resulting data keeps perception and caption content aligned while providing structured references for both SFT and RL. We evaluate PercepCap on caption benchmarks such as DREAM-1K~\cite{dream1k}, CaReBench~\cite{carebench} and VidCapBench-AE~\cite{vidcapbench} and caption-to-QA benchmarks like MotionBench~\cite{motionbench}, and ShortVidBench~\cite{shortvidbench}. The results show that our model achieves leading open-source caption quality. Our main contributions are as follows:

\begin{itemize}
    \item \textbf{Perception-aware captioning framework.} Before the final caption, we formulate video captioning with an explicit spatio-temporal perception trace, including object trajectories and temporally grounded events, turning captioning into evidence-grounded generation.
    \item \textbf{Perceive-then-Describe SFT and Perception-Grounded RL.} We train PercepCap in two stages. PD-SFT adapts the model to the \emph{perceive--describe} chain, while PG-RL optimizes trace grounding and final caption quality with decomposed, stage-specific perception-level and caption-level rewards.
    \item \textbf{Caption-Anchored Perception Data Construction.} We construct caption-aligned perception data by first generating a caption-only description, extracting its object and event content, and grounding the extracted content back in the video to generate boxes and timestamps. This construction aligns objects and events between the perception trace and final caption, supporting both PD-SFT and PG-RL.
\end{itemize}

\section{Related Work}
\label{sec:related_work}

\subsection{Video Captioning and Multi-modal Large Language Models}
Video captioning generates descriptive natural language for video sequences. Recent work based on Multi-modal Large Language Models (MLLMs)~\cite{lin2023video,zhang2024llavanext,bai2023qwen,qwen3.5,Qwen2.5-VL} has made the requirements of detailed video captioning more explicit. Tarsier~\cite{dream1k} introduces a family of video captioners and releases DREAM-1K, a challenging benchmark with an automatic evaluation method for fine-grained video caption. AuroraCap~\cite{vdc} studies efficient detailed caption generation with token merging and introduces VDC, a benchmark of structured detailed captions together with an LLM-assisted metric VDCscore. CaReBench~\cite{carebench} further targets fine-grained captioning with manually separated spatial and temporal annotations. These works follow a \emph{caption-only generation paradigm}, emitting the final caption without an explicit perception trace. PercepCap instead targets the generation process by producing a structured spatio-temporal perception trace of objects and events and conditioning the final caption on this evidence.

\subsection{Perception-Augmented Video Understanding}
Fine-grained video perception includes multiple low-level tasks such as multi-object tracking and temporal localization~\cite{DBLP:journals/corr/abs-2512-14698,DBLP:conf/cvpr/RenYL0H24}.
A growing trend in MLLMs seeks to unify perception and language generation by prompting models to generate spatial boxes or temporal timestamps directly as text.
A closely related work, VideoP2R~\cite{videop2r}, makes intermediate visual evidence explicit through a single free-form observation before producing the final answer for video question answering. PercepCap draws inspiration from this explicit-evidence design but adapts it to detailed video captioning. Video question answering is usually guided by a specific question, so the intermediate observation can focus on the evidence needed for that answer. Detailed captioning, in contrast, must describe the video more comprehensively: it needs to track which objects appear, how their locations change over time, and when different events start and end. Therefore, PercepCap organizes intermediate evidence as a structured perception trace, replacing the single free-form observation with spatial object identities and trajectories, together with temporal event descriptions and boundaries.

\subsection{Reinforcement Learning for MLLMs}

Reinforcement Learning from Human Feedback (RLHF)~\cite{ouyang2022rlhf} and RL-based post-training~\cite{DBLP:conf/acl/TrungZJSJL24} have been widely used to improve instruction following and align model behavior. In multimodal settings, RL has become an important direction for improving perception and reasoning. Group-based methods such as Group Relative Policy Optimization (GRPO)~\cite{DBLP:journals/corr/abs-2501-12948} estimate relative advantages from multiple responses sampled under the same prompt, thereby avoiding the need for an explicit value model.

Recent work has begun to extend RL-based post-training to video MLLMs, with a focus on spatio-temporal perception, multimodal reasoning, and, more recently, video captioning. VideoChat-R1~\cite{DBLP:journals/corr/abs-2504-06958} applies reinforcement fine-tuning to improve spatio-temporal perception in video MLLMs. VideoChat-R1.5~\cite{DBLP:journals/corr/abs-2509-21100} further introduces visual test-time scaling, which refines visual perception during inference to strengthen multimodal reasoning. VideoCap-R1~\cite{DBLP:journals/corr/abs-2506-01725} moves this direction toward video captioning by applying GRPO-based RL post-training to structured caption generation. VidBridge-R1~\cite{vidbridger1} studies RL across video QA and captioning, using intermediate proxy tasks to bridge the gap between convergent QA-style reasoning and divergent caption-style description.

Together, these studies suggest that RL-based post-training is a promising way to improve video perception, reasoning, and caption generation in MLLMs. PercepCap builds on this direction, but shifts the focus from rewarding only the final response or a coarse intermediate process to supervising explicit perceptual evidence. Specifically, it assigns decomposed rewards to object trajectories, event boundaries, and the final caption. This provides separate training signals for what the model perceives and how it verbalizes that perception, encouraging stronger alignment between perceived entities, perceived events, and the generated caption.

\section{PercepCap}
\label{sec:method}

\begin{figure}[t]
\centering
\includegraphics[width=0.96\linewidth]{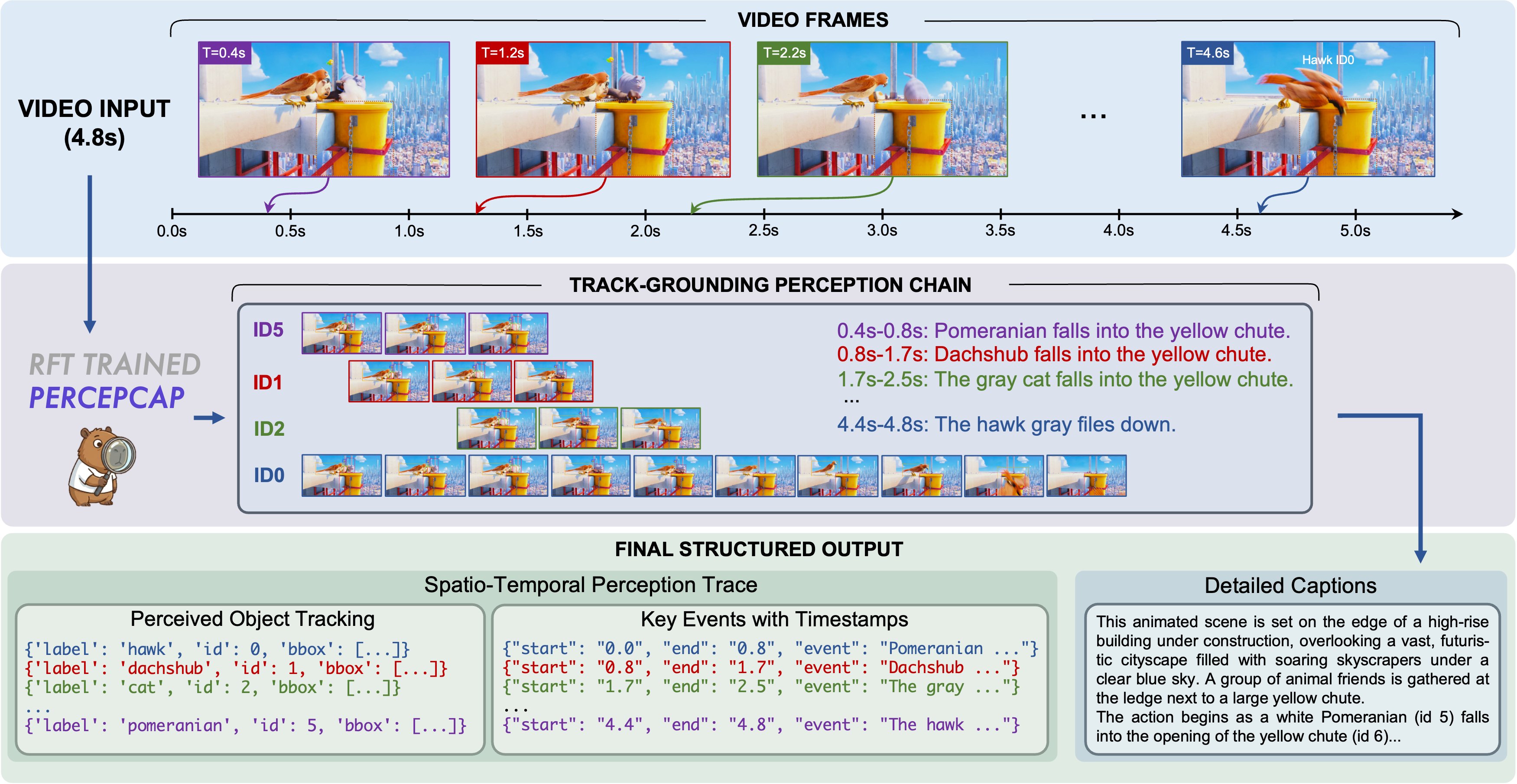}
\caption{Overview of PercepCap's \emph{perceive--describe} chain. The model first generates a structured spatio-temporal perception trace and then produces the final caption conditioned on this evidence.}
\label{fig:percepcap_framework}
\end{figure}

We present \textbf{PercepCap}, a perception-aware video captioning framework that explicitly separates \emph{spatio-temporal perception} from \emph{caption generation}. As illustrated in Figure~\ref{fig:percepcap_framework}, PercepCap follows a \emph{perceive--describe} chain: it first generates a structured \emph{perception trace} that captures object- and event-level evidence, and then produces the final caption conditioned on this trace. We train PercepCap with a two-stage pipeline. First, \textbf{Perceive-then-Describe Supervised Fine-tuning (PD-SFT)} teaches the model to generate the perception trace before producing the caption. Second, \textbf{Perception-Grounded Reinforcement Learning (PG-RL)} further improves the grounding of the perception trace and the quality of the final caption through perception-grounded rewards.

\subsection{Structured Spatio-Temporal Perception}

Given an input video $V = \{v_1, v_2, \dots, v_T\}$, the goal of video captioning is to generate a natural language description $Y_{\text{cap}}$ that is fluent and grounded in the visual content. PercepCap reformulates the standard caption-only target as a \textbf{perceive--describe output sequence}, denoted as $Y = [Y_{\text{perc}}, Y_{\text{cap}}]$, where $Y_{\text{perc}}$ is a structured spatio-temporal perception trace and $Y_{\text{cap}}$ is the final caption conditioned on this trace. The perception trace contains two complementary types of evidence: spatial object evidence and temporal event evidence. The serialized output format is provided in Appendix~\ref{app:structured_output}.

\textbf{Objects.} Detailed captions need to preserve not only which entities appear in the video, but also where they are located and how they move over time. We therefore represent objects as explicit spatial perception evidence. Each object entry contains a temporally consistent identity \texttt{id}, a semantic \texttt{label}, and a \texttt{localization} sequence. Each localization item records a \texttt{timestamp} and a 2D bounding box \texttt{bbox\_2d} specified by relative pixel coordinates, range [0, 1000].

\textbf{Events.} Detailed captions also require knowing what happens and when it happens, rather than merely listing visible entities. We therefore represent events as temporal perception evidence. Each event entry contains a temporal segment, specified by \texttt{start\_time} and \texttt{end\_time}, together with a concise textual \texttt{event} description. The subjects mentioned in event descriptions are encouraged to be consistent with the objects tracked in the \texttt{objects} field, thereby linking temporal actions to spatially grounded entities.

Overall, this \emph{perceive--describe} formulation turns detailed captioning into evidence-grounded generation. The final description is conditioned on explicit object trajectories and temporally grounded events, rather than being produced as a single undifferentiated caption. The structured trace also separates the output into interpretable components, which can be supervised during PD-SFT and further optimized with component-level rewards during PG-RL.

\subsection{Perceive-then-Describe Supervised Fine-tuning}

We use supervised fine-tuning to adapt the model from direct caption generation to the \emph{perceive--describe} output sequence defined above. Each training example consists of an input video $V$ and a ground-truth target sequence $Y^{\star}=[Y^{\star}_{\text{perc}},Y^{\star}_{\text{cap}}]$, where $Y^{\star}_{\text{perc}}$ is the structured perception trace and $Y^{\star}_{\text{cap}}$ is the final caption grounded in that trace. We optimize the \emph{next-token cross-entropy} over the serialized target sequence:
\begin{equation}
\mathcal{L}_{\text{SFT}} = -\frac{1}{|Y^{\star}|}\sum_{t=1}^{|Y^{\star}|} \log p_{\theta}(Y_t = Y^{\star}_t \mid V, Y^{\star}_{<t}).
\end{equation}

This stage serves two purposes. First, it teaches the model to follow the sequential generation chain, producing object trajectories and temporally grounded events before generating the final caption. Second, it makes the caption depend on an explicit perception trace, creating a structured output space in which trace grounding and caption quality can be separately optimized in the subsequent PG-RL stage.

\subsection{Perception-Grounded Reinforcement Learning}
\label{sec:rft}

Although PD-SFT teaches the model to follow the \emph{perceive--describe} chain, it optimizes the serialized perception trace and final caption with a single next-token objective. This coupled training signal cannot distinguish whether an error should be attributed to inaccurate perception, incomplete captioning, or a mismatch between the two. To address this limitation, PG-RL grounds reinforcement learning in the structured perception trace by combining perception-level and caption-level feedback in a composite reward.

\paragraph{Group Relative Policy Optimization.}
We adopt GRPO~\cite{DBLP:journals/corr/abs-2501-12948} as the RL optimizer. Given a training prompt $q$ consisting of a video and an instruction, the old policy samples a group of $G$ candidate outputs $\{o_i\}_{i=1}^G$, each scored by a scalar reward $r_i$. Following DeepSeek-R1~\cite{DBLP:journals/corr/abs-2501-12948}, GRPO optimizes the clipped group-relative objective
\begin{equation}
\begin{aligned}
\mathcal{J}_{\mathrm{GRPO}}(\theta)
=&~\mathbb{E}_{\substack{q\sim P(Q),\\
\{o_i\}_{i=1}^{G}\sim\pi_{\theta_{\mathrm{old}}}(O\mid q)}}
\Bigg[
\frac{1}{G}\sum_{i=1}^{G}
\bigg(
\min\!\left[
\rho_i(\theta) A_i,
\widetilde{\rho}_i(\theta) A_i
\right]
-\eta\,D_{\mathrm{KL}}\!\left(\pi_{\theta}\,\|\,\pi_{\mathrm{ref}}\right)
\bigg)
\Bigg],
\end{aligned}
\end{equation}
where
\begin{equation}
\begin{aligned}
\rho_i(\theta)=
\frac{\pi_{\theta}(o_i\mid q)}
{\pi_{\theta_{\mathrm{old}}}(o_i\mid q)},
\qquad
\widetilde{\rho}_i(\theta)=
\operatorname{clip}\!\left(\rho_i(\theta),1-\epsilon,1+\epsilon\right)
\end{aligned}
\end{equation}
are the output-level policy ratio and clipped ratio.
Rewards are normalized within the group to compute the relative advantage
\begin{equation}
A_i = \frac{r_i - \mathrm{mean}(\{r_j\}_{j=1}^G)}{\mathrm{std}(\{r_j\}_{j=1}^G)}.
\end{equation}

\paragraph{Perception-grounded rewards.}
PercepCap generates two output segments, $o = [o_{\text{perc}},\, o_{\text{cap}}]$, and optimizes them with a reward function that covers both the intermediate perception trace and the final caption. The reward consists of the following components:

1)~\emph{Perception-level spatial reward $R_{\mathrm{iou}}$.} To encourage spatially accurate object localization in $o_{\text{perc}}$, we use a GT-centric coverage IoU reward over the predicted \texttt{localization} entries. For each GT timestamp, predicted and ground-truth boxes with the same timestamp are paired by Hungarian matching, and the matched pairs are scored by their box IoU.

2)~\emph{Perception-level temporal rewards $R_{\mathrm{tiou}}$ and $R_{\mathrm{sodam}}$.} We use two rewards over predicted events in $o_{\text{perc}}$. $R_{\mathrm{tiou}}$ evaluates event timestamp quality: it computes pairwise temporal IoU between predicted and ground-truth intervals, finds an order-preserving alignment, merges adjacent predicted intervals assigned to the same ground-truth event, and then computes F1 from precision and recall averaged over multiple temporal-IoU thresholds. $R_{\mathrm{sodam}}$ reuses the same aligned event groups and uses an LLM judge to determine whether the grouped predicted event text is semantically consistent with the ground-truth event.

3)~\emph{Caption reward $R_{\mathrm{cap}}$.} We evaluate only the final caption segment $o_{\text{cap}}$. For objects and events, an LLM judge first checks whether objects and events extracted from the ground-truth caption are entailed by the generated caption, yielding recall-oriented scores. It then extracts object and event mentions from the generated caption and checks whether these predicted items are supported by the ground-truth caption, yielding precision-oriented scores. We convert the object and event precision--recall pairs into object F1 and event F1, and sum them as $R_{\mathrm{cap}}$. This rewards coverage of salient entities and events while penalizing unsupported caption content.

4)~\emph{Format reward $R_{\mathrm{format}}$.} To stabilize training and prevent reward computation failures caused by malformed outputs, we enforce a strict JSON-format constraint. The generated output must be parseable as the required structured output, contain the required \texttt{perception}, \texttt{objects}, \texttt{events}, and \texttt{description} fields, and provide well-formed object identities, timestamped boxes, event intervals, and a non-empty final description.

Finally, we score each output $o_i$ with a composite reward:
\begin{equation}
r_i = R(o_i) =
\underbrace{
\alpha R_{\mathrm{iou}}(o_i)
+ \beta R_{\mathrm{tiou}}(o_i)
+ \delta R_{\mathrm{sodam}}(o_i)
}_{\text{perception reward}}
+ \lambda R_{\mathrm{format}}(o_i)
+ \gamma R_{\mathrm{cap}}(o_i).
\end{equation}
The perception rewards evaluate the intermediate trace from three aspects: whether objects are localized accurately, whether event timestamps are correct, and whether event descriptions match the ground-truth events. In contrast, $R_{\mathrm{cap}}$ evaluates only the final caption, and $R_{\mathrm{format}}$ ensures that the structured output remains parseable. By assigning rewards to both the perception trace and the final caption, PG-RL avoids optimizing caption quality alone and better matches the \emph{perceive--describe} generation chain.

\section{Caption-Anchored Perception Data Construction}
\label{sec:data_construction}

Standard video-caption datasets usually provide only final captions, whereas PercepCap requires paired annotations for both the perception trace and the final description. We therefore construct \emph{caption-aligned perception data} for each training video:
$Y^{\star}=[Y^{\star}_{\text{perc}},Y^{\star}_{\text{cap}}]$.
Here, $Y^{\star}_{\text{cap}}$ is the reference caption generated by the annotation model, and $Y^{\star}_{\text{perc}}$ is a structured perception trace whose objects and events are aligned with the content of that caption. To construct such paired data with spatial object boxes and temporal event boundaries, we propose \emph{Caption-Anchored Perception Data Construction}. As illustrated in Figure~\ref{fig:data_construction_example}, the process consists of four steps:

\begin{figure}[t]
\centering
\includegraphics[width=\linewidth]{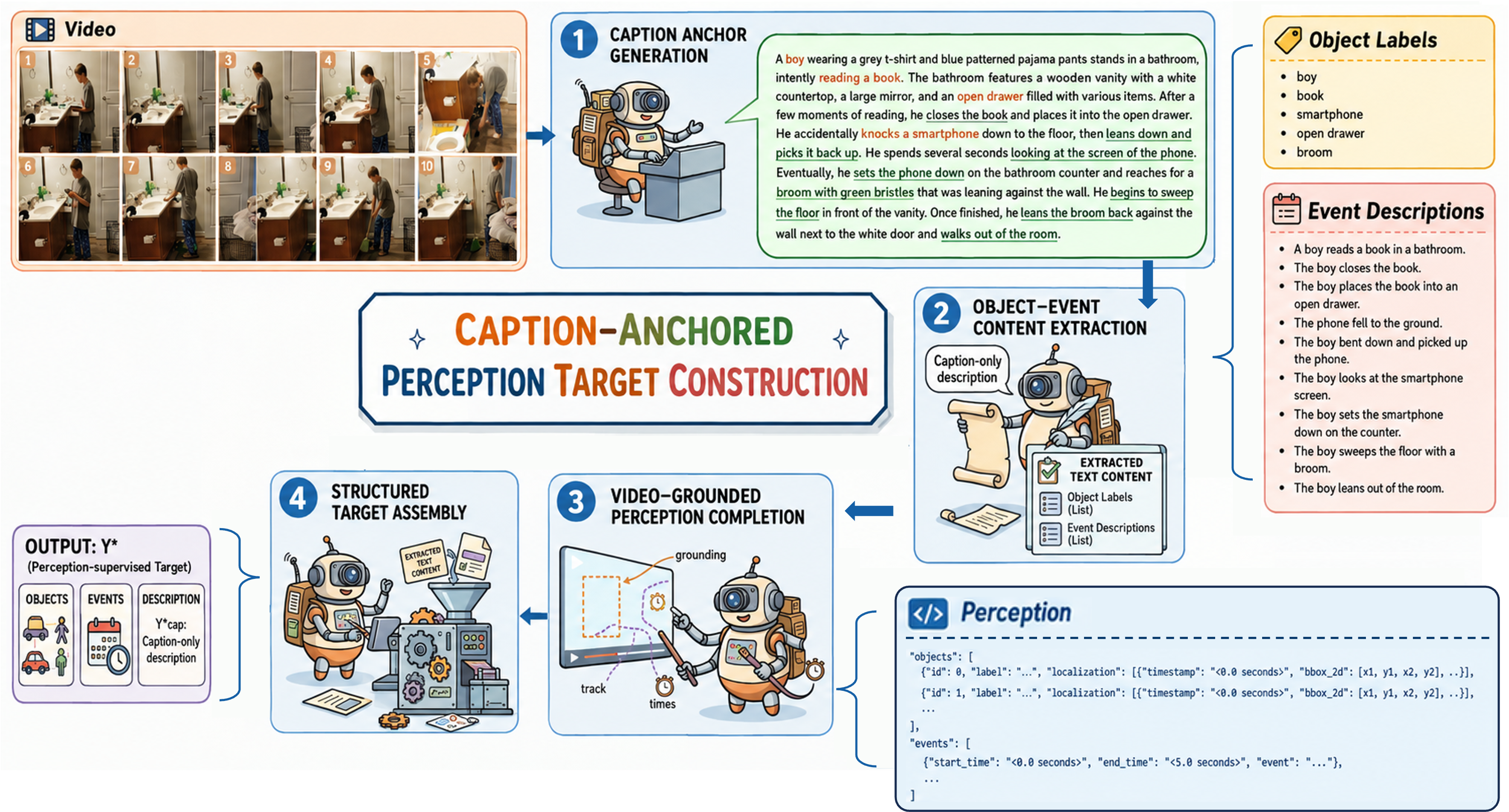}
\caption{Example and construction flow for Caption-Anchored Perception Data Construction. A caption-only description anchors the object and event content, which is then grounded back in the video and assembled into caption-aligned perception data.}
\label{fig:data_construction_example}
\end{figure}

1)~\emph{Caption Anchor Generation.} Given a training video, we use a strong video model as the annotation model, under the caption-only setting to produce $Y^{\star}_{\text{cap}}$. The annotation model, outputs the caption without perception, which serves as the semantic anchor for subsequent perception construction.

2)~\emph{Object-Event Content Extraction.} To make the trace cover the same objects and events as the final description, we ask the annotation model, to extract from the caption the textual content to be grounded, including object labels and event descriptions. Object labels specify entities in the trace, while event descriptions specify actions or temporal changes that need boundaries.

3)~\emph{Video-Grounded Perception Completion.} Next, the annotation model revisits the video and completes the extracted content with grounding information. For each object label, it assigns identity and predicts timestamped boxes; for each event description, it predicts start and end times. This step completes missing spatial and temporal fields for the extracted content.

4)~\emph{Structured Data Assembly.} Finally, we assemble the completed object tracks, event segments, and original caption into the structured output format in Section~\ref{sec:method}. The resulting $Y^{\star}$ contains \texttt{objects}, \texttt{events}, and \texttt{description}; the description is the caption-only output from the annotation model, and the perception fields are grounded completions of its extracted objects and events. The data provides PD-SFT supervision and references for PG-RL rewards.

Overall, the caption-anchored construction \emph{strictly enforces object-event consistency} between the perception trace and the final caption. The caption determines which objects and events should be described, and video-grounded completion supplies where they appear and when they occur. This data supports both PD-SFT and PG-RL.

\section{Results and Analysis}
\label{sec:results}

\begin{table}[t]
\centering
\begin{threeparttable}
\caption{\textbf{Evaluation results on DREAM-1K~\cite{dream1k}, CaReBench~\cite{carebench}, ShortVidBench~\cite{shortvidbench}, and MotionBench~\cite{motionbench}.} We highlight the \textbf{best} results in bold and \underline{second-best} results with underlining among open-source models.}
\label{tab:main_results}
\scriptsize
\setlength{\tabcolsep}{2pt}
\begin{tabular}{@{}
>{\raggedright\arraybackslash}m{0.19\linewidth}
>{\centering\arraybackslash}m{0.15\linewidth}|
>{\centering\arraybackslash}m{0.16\linewidth}
>{\centering\arraybackslash}m{0.16\linewidth}|
>{\centering\arraybackslash}m{0.13\linewidth}|
>{\centering\arraybackslash}m{0.13\linewidth}
@{}}
\toprule
\multicolumn{1}{c}{\multirow{2}{*}{Model}} & \multicolumn{1}{c|}{DREAM-1K~\cite{dream1k}} & \multicolumn{2}{c|}{CaReBench~\cite{carebench}} & \multicolumn{1}{c|}{ShortVidBench~\cite{shortvidbench}} & MotionBench~\cite{motionbench} \\
\cmidrule(l){2-6}
& \multicolumn{1}{c|}{Event F1/R/P} & Action F1/R/P & \multicolumn{1}{c|}{Object F1/R/P} & \multicolumn{1}{c|}{Acc.} & Acc. \\
\midrule
\textit{Proprietary models} & & & & & \\
{\color[HTML]{9B9B9B} Gemini 3 Flash~\cite{gemini3flash}} & \multicolumn{1}{c|}{{\color[HTML]{9B9B9B} 37.6/48.5/30.7}} & {\color[HTML]{9B9B9B} 42.4/39.6/45.7} & \multicolumn{1}{c|}{{\color[HTML]{9B9B9B} 40.3/37.8/43.2}} & \multicolumn{1}{c|}{{\color[HTML]{9B9B9B} 78.6}} & {\color[HTML]{9B9B9B} 57.6} \\
{\color[HTML]{9B9B9B} Gemini 3.1 Pro~\cite{geminiteam2026gemini31problog}} & \multicolumn{1}{c|}{{\color[HTML]{9B9B9B} 38.2/45.2/33.1}} & {\color[HTML]{9B9B9B} 42.0/37.4/47.7} & \multicolumn{1}{c|}{{\color[HTML]{9B9B9B} 39.3/32.6/49.6}} & \multicolumn{1}{c|}{{\color[HTML]{9B9B9B} 78.4}} & {\color[HTML]{9B9B9B} 55.9} \\
\midrule
\textit{Open-source models} & & & & & \\
Qwen2.5-VL 7B~\cite{Qwen2.5-VL} & \multicolumn{1}{c|}{27.2/27.9/26.6} & 30.9/23.6/45.0 & \multicolumn{1}{c|}{34.1/27.6/44.6} & \multicolumn{1}{c|}{70.0} & 50.7 \\
Kimi-VL A3B~\cite{kimivl} & \multicolumn{1}{c|}{27.5/32.6/23.8} & 35.1/29.0/44.4 & \multicolumn{1}{c|}{37.7/32.0/\textbf{45.8}} & \multicolumn{1}{c|}{74.2} & 52.1 \\
GLM 4.6v 10B~\cite{glm45v} & \multicolumn{1}{c|}{30.2/32.5/\underline{28.3}} & 34.2/27.3/\textbf{45.9} & \multicolumn{1}{c|}{\underline{39.7}/\underline{36.3}/43.9} & \multicolumn{1}{c|}{72.4} & 51.6 \\
Qwen3-VL 8B~\cite{qwen3technicalreport} & \multicolumn{1}{c|}{29.1/36.3/24.3} & 35.9/31.2/42.2 & \multicolumn{1}{c|}{38.6/35.4/42.5} & \multicolumn{1}{c|}{72.8} & 53.0 \\
Qwen3.5 9B~\cite{qwen3.5} & \multicolumn{1}{c|}{\underline{30.3}/\underline{40.5}/24.1} & \underline{36.9}/\textbf{34.8}/39.3 & \multicolumn{1}{c|}{38.3/\textbf{42.1}/35.1} & \multicolumn{1}{c|}{\underline{74.8}} & \textbf{56.6} \\
\midrule
\rowcolor[HTML]{EFEFEF}
\cellcolor[HTML]{EFEFEF}\textbf{PercepCap (Ours)} & \multicolumn{1}{c|}{\cellcolor[HTML]{EFEFEF}\textbf{33.9}/\textbf{41.4}/\textbf{28.7}} & \textbf{39.4}/\underline{34.7}/\underline{45.6} & \multicolumn{1}{c|}{\cellcolor[HTML]{EFEFEF}\textbf{40.4}/\underline{36.3}/\underline{45.4}} & \multicolumn{1}{c|}{\cellcolor[HTML]{EFEFEF}\textbf{76.1}} & \underline{54.9} \\
\bottomrule
\end{tabular}
\end{threeparttable}
\vspace{-3mm}
\end{table}

\begin{table}[t]
\centering
\caption{\textbf{Evaluation results on VidCapBench-AE~\cite{vidcapbench}.} We highlight the \textbf{best} results in bold and \underline{second-best} results with underlining among open-source models.}
\label{tab:vidcapbench_ae_results}
\scriptsize
\setlength{\tabcolsep}{3pt}
\resizebox{\linewidth}{!}{
\begin{tabular}{@{}lccccc@{}}
\toprule
\multicolumn{1}{c}{\multirow{3}{*}{Model}} & \multicolumn{5}{c}{VidCapBench-AE~\cite{vidcapbench}} \\
\cmidrule(l){2-6}
& \multicolumn{1}{c|}{Overall} & \multicolumn{1}{c|}{Video Aesthetics} & \multicolumn{1}{c|}{Video Content} & \multicolumn{1}{c|}{Video Motion} & Physical Laws \\
\cmidrule(l){2-6}
& \multicolumn{1}{c|}{Acc./Pre./Cov./Con.} & \multicolumn{1}{c|}{Acc./Pre./Cov./Con.} & \multicolumn{1}{c|}{Acc./Pre./Cov./Con.} & \multicolumn{1}{c|}{Acc./Pre./Cov./Con.} & Acc./Pre./Cov./Con. \\
\midrule
\textit{Proprietary models} & & & & & \\
{\color[HTML]{9B9B9B} GPT-4o} & \multicolumn{1}{c|}{{\color[HTML]{9B9B9B} 16.8/57.4/86.0/5.9}} & \multicolumn{1}{c|}{{\color[HTML]{9B9B9B} 14.1/47.6/83.4/4.9}} & \multicolumn{1}{c|}{{\color[HTML]{9B9B9B} 17.5/61.7/87.2/6.1}} & \multicolumn{1}{c|}{{\color[HTML]{9B9B9B} 10.2/41.3/84.0/3.6}} & {\color[HTML]{9B9B9B} 27.9/62.1/85.4/9.7} \\
{\color[HTML]{9B9B9B} Gemini 3 Flash~\cite{gemini3flash}} & \multicolumn{1}{c|}{{\color[HTML]{9B9B9B} 19.3/58.1/90.5/7.3}} & \multicolumn{1}{c|}{{\color[HTML]{9B9B9B} 17.1/47.3/87.5/6.4}} & \multicolumn{1}{c|}{{\color[HTML]{9B9B9B} 19.6/62.8/91.7/7.4}} & \multicolumn{1}{c|}{{\color[HTML]{9B9B9B} 12.9/49.5/89.8/4.9}} & {\color[HTML]{9B9B9B} 33.9/58.4/91.7/12.8} \\
\midrule
\textit{Open-source models} & & & & & \\
Qwen2-VL 7B & \multicolumn{1}{c|}{11.1/47.1/77.0/\underline{6.4}} & \multicolumn{1}{c|}{12.4/44.3/78.7/\textbf{7.2}} & \multicolumn{1}{c|}{9.9/48.3/75.9/5.7} & \multicolumn{1}{c|}{4.0/22.7/78.2/2.3} & 26.1/\textbf{59.4}/81.2/\textbf{15.1} \\
Kimi-VL A3B~\cite{kimivl} & \multicolumn{1}{c|}{15.4/54.5/85.9/2.8} & \multicolumn{1}{c|}{13.5/46.9/81.1/2.4} & \multicolumn{1}{c|}{15.7/58.3/87.7/2.8} & \multicolumn{1}{c|}{6.2/36.5/85.3/1.1} & 28.1/52.0/\underline{89.4}/5.1 \\
MiMo-VL~\cite{mimovl} & \multicolumn{1}{c|}{15.8/52.3/86.8/\textbf{7.5}} & \multicolumn{1}{c|}{14.5/44.9/85.0/\underline{6.9}} & \multicolumn{1}{c|}{15.8/55.8/87.8/\textbf{7.5}} & \multicolumn{1}{c|}{8.4/36.1/84.9/\textbf{4.0}} & 28.4/56.8/85.4/\underline{13.6} \\
Intern2VL 8B~\cite{intern2vl} & \multicolumn{1}{c|}{10.2/43.0/84.9/2.5} & \multicolumn{1}{c|}{9.1/36.3/84.4/2.2} & \multicolumn{1}{c|}{10.0/46.1/85.2/2.4} & \multicolumn{1}{c|}{4.4/18.0/81.3/1.1} & 23.6/52.8/85.7/5.8 \\
Qwen3-VL 8B~\cite{qwen3technicalreport} & \multicolumn{1}{c|}{\underline{17.8}/\underline{56.8}/\underline{88.5}/4.4} & \multicolumn{1}{c|}{14.5/\underline{47.3}/84.8/3.6} & \multicolumn{1}{c|}{\underline{18.6}/\underline{61.1}/\underline{90.3}/4.6} & \multicolumn{1}{c|}{\underline{9.3}/\underline{39.2}/84.0/2.3} & \underline{32.7}/\underline{57.3}/87.7/8.1 \\
Qwen3.5 9B~\cite{qwen3.5} & \multicolumn{1}{c|}{17.2/52.7/\underline{88.5}/5.8} & \multicolumn{1}{c|}{\underline{15.3}/44.5/\underline{85.5}/5.2} & \multicolumn{1}{c|}{17.4/56.7/89.6/5.9} & \multicolumn{1}{c|}{\underline{9.3}/37.6/\textbf{91.1}/3.2} & 31.2/52.2/\underline{89.4}/10.5 \\
\midrule
\rowcolor[HTML]{EFEFEF}
\cellcolor[HTML]{EFEFEF}\textbf{PercepCap (Ours)} & \multicolumn{1}{c|}{\cellcolor[HTML]{EFEFEF}\textbf{19.0}/\textbf{59.1}/\textbf{91.0}/6.3} & \multicolumn{1}{c|}{\cellcolor[HTML]{EFEFEF}\textbf{16.2}/\textbf{47.6}/\textbf{88.8}/5.1} & \multicolumn{1}{c|}{\cellcolor[HTML]{EFEFEF}\textbf{19.5}/\textbf{64.7}/\textbf{92.1}/\underline{6.6}} & \multicolumn{1}{c|}{\cellcolor[HTML]{EFEFEF}\textbf{9.8}/\textbf{41.1}/\underline{87.6}/\underline{3.5}} & \textbf{34.9}/56.6/\textbf{92.0}/10.7 \\
\bottomrule
\end{tabular}

}
\vspace{-3mm}
\end{table}

\subsection{Implementation Details}
\label{sec:implementation_details}
We use the open-source Qwen3-VL-8B-Instruct~\cite{qwen3technicalreport} checkpoint under the caption-only generation paradigm as the starting comparison: given the video input, the model is prompted to generate only the final caption without producing an explicit perception trace. PercepCap keeps the same model architecture and introduces no additional modules. We construct caption-aligned perception data on approximately 57K action-rich videos of 15--30 seconds, where caption anchors are generated by Gemini~3 Flash~\cite{gemini3flash} and the structured perception fields are completed by Gemini~3.1 Pro~\cite{geminiteam2026gemini31problog}. Starting from the Qwen3-VL checkpoint, we improve caption generation through training and a perception-aware generation chain. PD-SFT on the constructed data adapts the model to the \emph{perceive--describe} chain, where it produces structured spatio-temporal perception before the caption. PG-RL, implemented with GRPO~\cite{DBLP:journals/corr/abs-2501-12948}, further optimizes the same model with the reward design described in \ref{sec:rft}. During RL, we optimize with AdamW and set the reward coefficients to $\alpha=\beta=\delta=0.5$ and $\gamma=\lambda=1.0$ to balance perception-level and caption-level objectives. The whole hyperparameter settings and system prompts are provided in Appendix~\ref{app:training_details}.

\subsection{Benchmarks and Evaluation Metrics}
\label{sec:experiments}

We evaluate PercepCap on four groups of detailed video captioning and video description benchmarks, following the official evaluation procedure for each benchmark. For LLM-judge-based evaluation, DREAM-1K~\cite{dream1k} and CaReBench~\cite{carebench} use GPT-4.1 as the judge because their original evaluation APIs are no longer available; as demonstrated in CaReBench~\cite{carebench}, the choice of LLM judge does not affect the validity of the comparison. ShortVidBench~\cite{shortvidbench} and MotionBench~\cite{motionbench} use Gemini~3.1 Pro as the judge, and VidCapBench-AE~\cite{vidcapbench} uses GPT-4o as in the original paper.
\textbf{1) DREAM-1K}~\cite{dream1k}: a benchmark for fine-grained video description with the AutoDQ evaluation procedure, for which we report \emph{F1/Precision/Recall}.
\textbf{2) CaReBench}~\cite{carebench}: a fine-grained benchmark for video captioning and retrieval with the CapST evaluation procedure, for which we report \emph{Action} and \emph{Object} \emph{F1/Recall/Precision}.
\textbf{3) ShortVidBench}~\cite{shortvidbench} \textbf{and MotionBench}~\cite{motionbench}: two complementary benchmarks for short-video and motion-oriented description, evaluated with a caption-to-QA procedure. An LLM judge answers the benchmark questions using the predicted caption as the evidence source, and the resulting QA accuracy reflects whether the caption preserves the information needed to answer questions about the original video. We report this \emph{Accuracy}.
\textbf{4) VidCapBench-AE}~\cite{vidcapbench}: a detailed video captioning benchmark that evaluates overall quality and category-specific aspects with GPT-4o as the judge, for which we report \emph{Accuracy/Precision/Coverage/Conciseness}.

\subsection{Main Results}
\label{sec:main_results}
We report results using the official metrics of each benchmark. Tables~\ref{tab:main_results} and \ref{tab:vidcapbench_ae_results} summarize the main comparisons on DREAM-1K~\cite{dream1k}, CaReBench~\cite{carebench}, ShortVidBench~\cite{shortvidbench}/MotionBench~\cite{motionbench}, and VidCapBench-AE~\cite{vidcapbench}, respectively. Across these evaluations, PercepCap shows consistent gains over the caption-only Qwen3-VL starting point, indicating that structured perception traces improve detailed video captioning beyond direct caption generation. On DREAM-1K~\cite{dream1k}, the higher AutoDQ F1 indicates better coverage of salient description units while maintaining competitive precision. On CaReBench~\cite{carebench}, the action and object scores show that PercepCap better preserves the entities and events needed for caption grounding. ShortVidBench~\cite{shortvidbench} and MotionBench~\cite{motionbench} further test whether the method generalizes to compact short-video and motion-focused evaluation settings, while VidCapBench-AE~\cite{vidcapbench} provides a more fine-grained view of overall quality, visual content, motion, aesthetics, and physical-law consistency. Taken together, these results support the effectiveness of PD-SFT and PG-RL for improving object/action coverage and grounded video description.

\begin{table}[t]
\caption{Ablation study on DREAM-1K measured by AutoDQ. Caption-only variants do not generate a perception trace, while perception-aware variants use the structured trace template.}
\label{tab:ablation}
\centering
\small
\begin{tabular}{@{}>{\centering\arraybackslash}p{0.05\textwidth}>{\raggedright\arraybackslash}p{0.49\textwidth}>{\centering\arraybackslash}p{0.103\textwidth}>{\centering\arraybackslash}p{0.103\textwidth}>{\centering\arraybackslash}p{0.103\textwidth}@{}}
\toprule
\# & Model Variant & F1 & Precision & Recall \\
\midrule
\#1 & Baseline & 29.1 & 36.3 & 24.3 \\
\#2 & + Perception chain from Gemini~3~Flash~\cite{gemini3flash} & 33.5 & 43.6 & 27.2 \\
\midrule
\#3 & + Caption-only SFT & 31.4 & 37.9 & 28.6 \\
\#4 & + Caption-only SFT + $R_{\text{cap}}$ only RL & 32.7 & 37.9 & 28.9 \\
\midrule
\#5 & + PD-SFT & 32.3 & 38.5 & 27.9 \\
\#6 & + PD-SFT + $R_{\text{cap}}$ only RL & 33.1 & 42.4 & 27.1 \\
\midrule
\rowcolor{gray!10}
\#7 & PercepCap (+ PD-SFT + PG-RL) & 33.9 & 41.4 & 28.7 \\
\bottomrule
\end{tabular}

\end{table}

\subsection{Ablation Study}
\label{sec:ablation}

\textbf{Effect of staged perception-aware training.} Starting from the Qwen3-VL caption-only setting in Table~\ref{tab:ablation} \#1, directly conditioning the same caption generator on PercepCap perception evidence (\#2) already improves AutoDQ in a zero-shot inference setting, showing that explicit perception can benefit captioning even before parameter updates. PD-SFT (\#5) then teaches the model to generate and use its own structured trace, while caption-level RL on top of this chain (\#6) separates final-caption optimization from the full joint reward suite. The final PercepCap setting (\#7) yields the best F1, showing that the staged perceive--describe pipeline benefits from both perception-level grounding signals and caption-level coverage signals.

\textbf{Perception-aware training versus caption-only training.} To verify that the improvement is not simply due to stronger final-caption supervision or additional fine-tuning, we compare matched caption-only and perception-aware variants in Table~\ref{tab:ablation}. The caption-only variants (\#3 and \#4) are trained with the same final captions but never generate a structured perception trace during training or inference; the RL variant (\#4) optimizes only final-caption rewards. Comparing \#3 with \#5 isolates the effect of adding the \emph{perceive--describe} chain during SFT, while comparing \#4 with \#6 isolates the effect of keeping the same caption-level RL objective but changing the generation paradigm from caption-only to perception-aware. The consistent gap between these paired settings suggests that the benefit comes from using perception as an explicit intermediate interface for both training and inference, rather than only from stronger caption data or more optimization.

\begin{figure}[t]
\centering
\includegraphics[width=\linewidth]{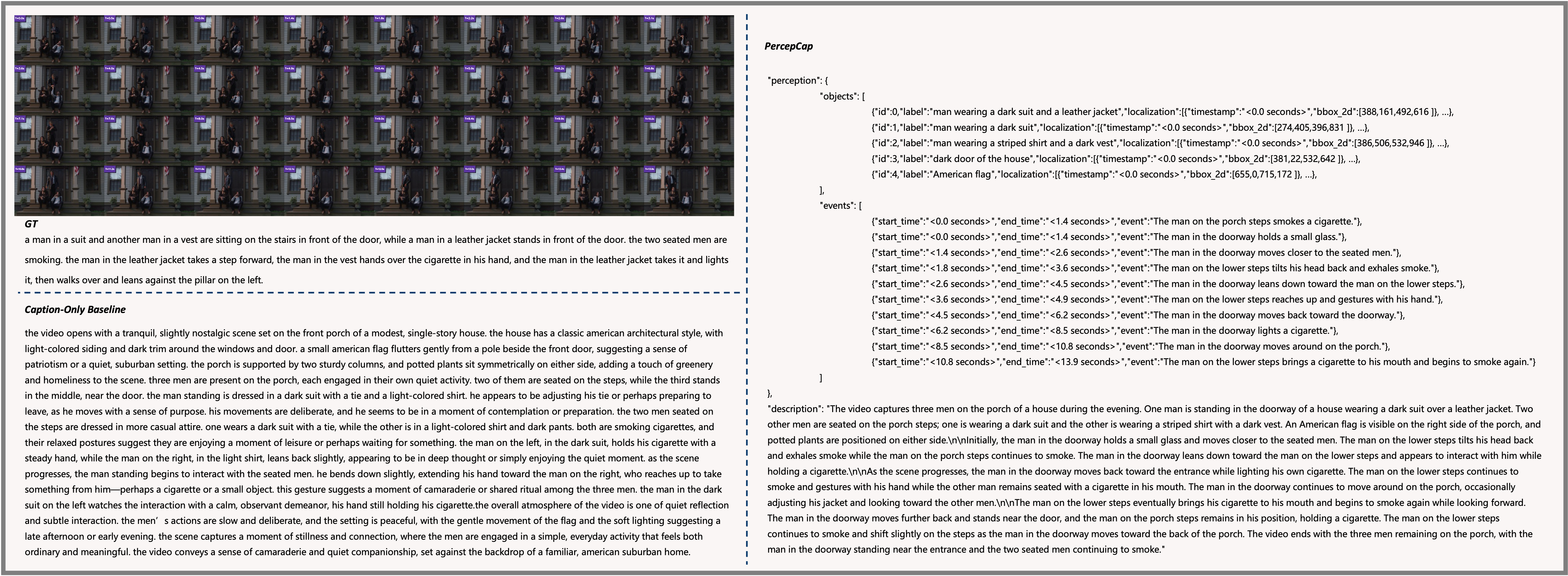}
\caption{Qualitative comparison between PercepCap and a direct single-step captioning baseline.}
\label{fig:qualitative}
\end{figure}

\textbf{Qualitative comparison.} In addition to the quantitative ablations above, Figure~\ref{fig:qualitative} compares captions from the caption-only baseline and PercepCap on the same video. PercepCap better distinguishes the key people in the scene and follows the cigarette handoff and lighting interaction, covering the seated men, the standing man, and their actions more faithfully. This leads to a caption with richer object coverage and a more complete action chain than the caption-only baseline.

\subsection{Limitations}
\label{sec:limitations}
PercepCap has several limitations. First, its supervision depends on annotator-generated perception data rather than manually verified object tracks and event boundaries, so errors in the constructed traces can propagate to SFT and RL references. Second, the \emph{perceive--describe} chain increases inference cost and output length because the model must generate a structured trace before the final caption. Third, the reward pipeline relies on parsing structured outputs and computing perception-level matching scores, making training more involved than caption-only optimization. Finally, the fixed trace schema improves auditability but may be less flexible for videos with dense interactions, long temporal horizons, or domain-specific objects and events.

\section{Conclusion}
\label{sec:conclusion}

We presented \textbf{PercepCap}, a perception-aware video captioning framework that turns direct caption generation into a \emph{perceive--describe} pipeline by making object trajectories and temporally grounded events explicit before the final caption. Since standard caption data lacks process-level perception supervision, Caption-Anchored Perception Data Construction first fixes the caption content and then grounds the same objects and events in the video, yielding caption-aligned perception data for training this pipeline. Perceive-then-Describe SFT teaches the model to follow this structured generation chain using this data, while Perception-Grounded RL assigns decomposed rewards to perception and caption segments to improve grounding, coverage, and credit assignment. Experiments on DREAM-1K, CaReBench, ShortVidBench, MotionBench, and VidCapBench-AE, together with ablation analysis, show that the staged training improves both detailed caption quality and explicit spatio-temporal perception. These results support structured perception traces as a practical intermediate interface for making caption evidence more inspectable and video captioning more auditable and grounded.

\newpage

{
    \small
    \bibliographystyle{ieeenat_fullname}
    \bibliography{neurips_2026}
}


\newpage
\appendix
\section{Supplementary Overview}
\label{sec:appendix}

This appendix collects implementation and evaluation details that support the main paper without changing its core claims. The main text is intended to stand alone; the material below documents data construction, prompts, training settings, and additional analysis procedures needed for reproducibility.

\section{Structured Perception Output Format}
\label{app:structured_output}

This section gives the concrete serialized format used for PercepCap's \emph{perceive--describe} output. The format instantiates the sequence $Y=[Y_{\mathrm{perc}},Y_{\mathrm{cap}}]$: the \texttt{perception} field stores the structured object and event evidence, and the \texttt{description} field stores the final caption conditioned on that evidence. This shared format is used during PD-SFT, perception-aware inference, output parsing, and reward computation.

\begin{center}
\begin{tcolorbox}[width=0.95\linewidth,colback=white,colframe=black,boxrule=0.4pt,arc=0pt,left=3pt,right=3pt,top=2pt,bottom=2pt]
\fontsize{6.7pt}{7.3pt}\selectfont\ttfamily
\{\\
\hspace*{1em}"perception": \{\\
\hspace*{2em}"objects": [\\
\hspace*{2em}\{\\
\hspace*{3em}"id": 0, "label": "...",\\
\hspace*{3em}"localization": [\\
\hspace*{4em}\{"timestamp": "<0.0 seconds>", "bbox\_2d": [x1, y1, x2, y2]\},\\
\hspace*{4em}\{"timestamp": "<0.2 seconds>", "bbox\_2d": [x1, y1, x2, y2]\},\\
\hspace*{4em}...\\
\hspace*{3em}]\\
\hspace*{2em}\},\\
\hspace*{2em}\{"id": 1, "label": "...", "localization": [...]\},\\
\hspace*{2em}...\\
\hspace*{2em}],\\
\hspace*{2em}"events": [\\
\hspace*{3em}\{"start\_time": "<0.0 seconds>", "end\_time": "<5.0 seconds>", "event": "..."\},\\
\hspace*{3em}...\\
\hspace*{2em}]\\
\hspace*{1em}\},\\
\hspace*{1em}"description": "..."\\
\}
\end{tcolorbox}
\end{center}

\section{Training and Inference Details}
\label{app:training_details}

PercepCap uses two post-training stages on the same Qwen3-VL backbone. During training, each video is sampled with a fixed 32 frames. PD-SFT trains the model to follow the \emph{perceive--describe} chain using the caption-anchored data from Section~\ref{app:data_construction}. We train PD-SFT for 2 epochs with a global batch size of 128, a learning rate of $2\times10^{-5}$, and a warmup ratio of 0.05.

PG-RL starts from the SFT checkpoint and optimizes the same structured output with GRPO and the joint rewards described in Section~\ref{sec:rft}. We use a global batch size of 256, prompt batch size of 16, generation batch size of 16, learning rate of $4\times10^{-6}$, GRPO group size $G=16$, and KL coefficient $\eta=0.001$. The reward coefficients are fixed to $\alpha=\beta=\delta=0.5$ and $\gamma=\lambda=1.0$.

At inference time, we use two prompt regimes. The caption-only baseline is prompted to generate only the final detailed caption, while PercepCap is prompted to generate a structured perception trace before the final caption. For benchmark scoring, we extract only the final caption; the perception trace is retained for auditing and optional perception-level evaluation. The exact inference prompts are shown in Section~\ref{app:prompt_details}. Videos are sampled with fixed 32 frames for all the models in DREAM-1K, CaReBench, ShortVidBench and MotionBench.

\section{Additional Qualitative Comparison}
\label{app:qualitative}

\begin{figure}[t]
\centering
\includegraphics[width=\linewidth]{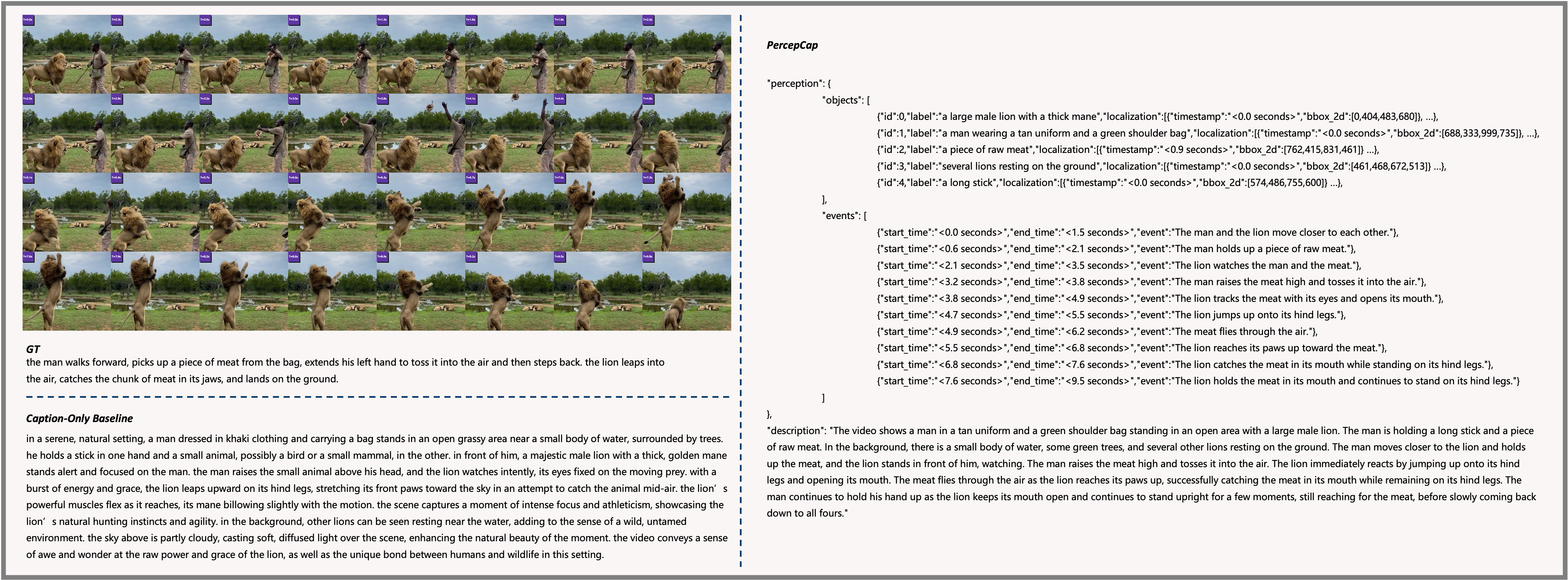}
\caption{Additional qualitative comparison between PercepCap and a direct single-step captioning baseline.}
\label{fig:appendix_qualitative_lion}
\end{figure}

Figure~\ref{fig:appendix_qualitative_lion} presents an additional DREAM-1K comparison. In this example, the caption-only baseline recognizes the man, bag, lion, and background scene, but misidentifies the thrown object as a small animal. PercepCap instead identifies the object as raw meat and keeps the action chain aligned around that object, including the man raising and tossing the meat, the lion opening its mouth, catching it, and returning to the ground. This example shows how more accurate object perception can support a more faithful event description in the final caption.

\section{Perception Evaluation}
\label{app:perception_evaluation}

We evaluate the perception ability of two representative variants from Table~\ref{tab:ablation}: the caption-only baseline (\#1) and the full PercepCap model (\#7). For this diagnostic evaluation, both variants are prompted to produce the same structured perception fields. For spatial grounding, we use YouTube-VOS~\cite{youtubevos}, a large-scale video object segmentation benchmark with dense object annotations, and derive reference object boxes from its annotated masks. We report the mean bbox IoU over frames where the predicted and ground-truth tracks both provide boxes, using an IoU threshold of 0.1 for boxes that participate in the metric. For temporal event grounding, we use ActivityNet~\cite{caba2015activitynet} as the evaluation set and report SODA\_m F1 following the TimeChat-Captioner event matching protocol~\cite{timechatcaptioner}: predicted event segments are aligned to ground-truth events with the order-aware DP matching procedure described in Section~\ref{sec:rft}, and the caption-similarity matching component is evaluated with Gemini~3 Flash. Table~\ref{tab:appendix_perception_eval} also provides AutoDQ caption metrics together with perception metrics. The results improve from \#1 to \#7 on both sides, indicating that PercepCap improves caption quality and explicit perception quality in tandem.

\begin{table}[t]
\centering
\caption{Caption and perception evaluation for representative ablation variants.}
\label{tab:appendix_perception_eval}
\scriptsize
\setlength{\tabcolsep}{3pt}
\resizebox{\linewidth}{!}{
\begin{tabular}{clccccc}
\toprule
\# & Model Variant & F1 & Precision & Recall & YouTube-VOS BBox IoU & ActivityNet SODA\_m F1 \\
\midrule
\#1 & Baseline & 29.1 & 36.3 & 24.3 & 48.42 & 33.84 \\
\#7 & PercepCap & 33.9 & 41.4 & 28.7 & 56.25 & 41.1 \\
\bottomrule
\end{tabular}
}
\end{table}

\section{Perception-Trace Data Construction}
\label{app:data_construction}

The SFT corpus is constructed by using an annotator model to produce caption-aligned perception data for each training video. Unlike a direct trace-generation pipeline, the construction first fixes the caption content and then grounds the same objects and events back into the video, which keeps the perception trace and final caption semantically aligned.

\begin{figure}[t]
\centering
\includegraphics[width=0.96\linewidth]{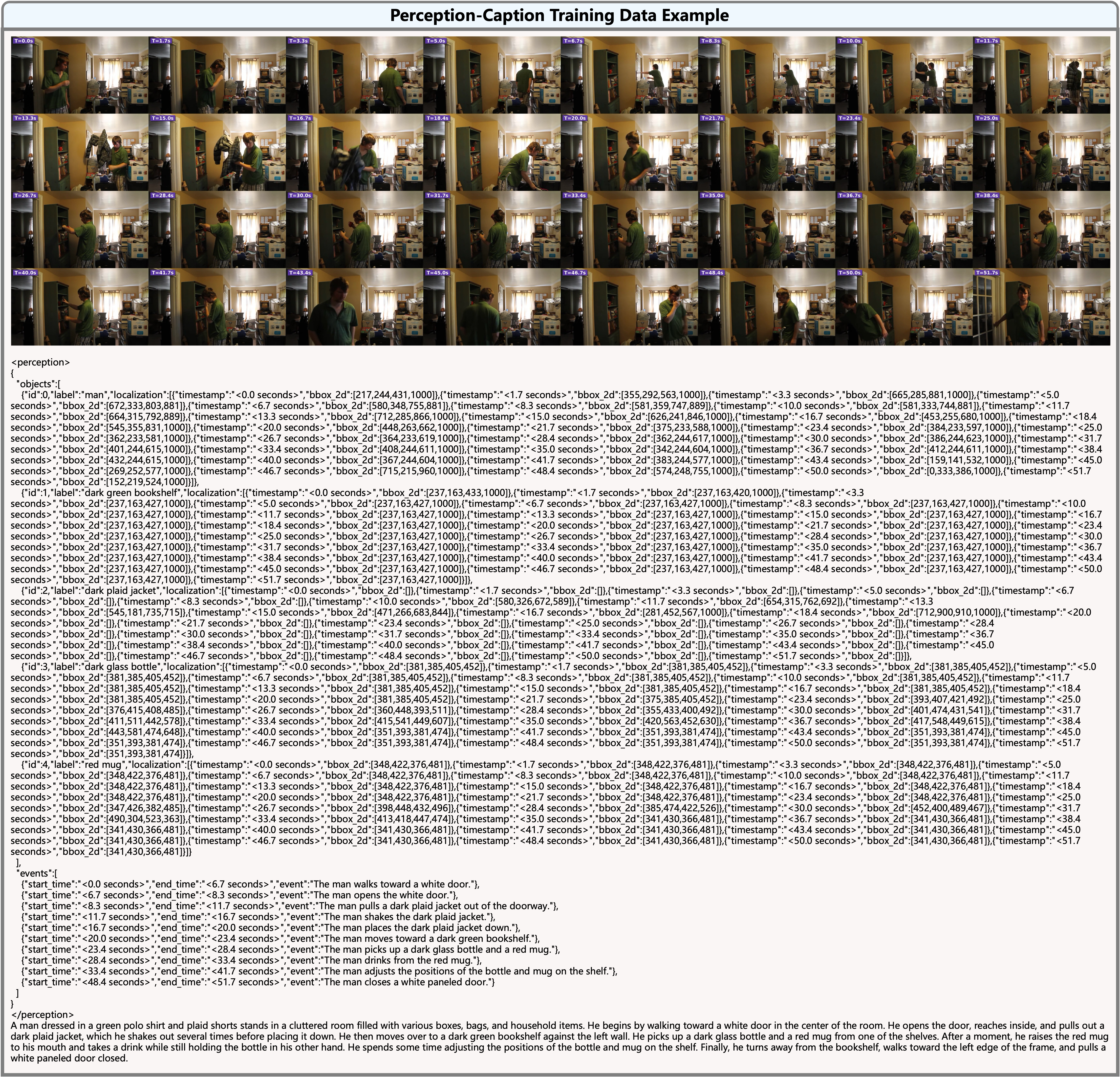}
\caption{Constructed training example from Caption-Anchored Perception Data Construction.}
\label{fig:data_example_1}
\end{figure}

\begin{figure}[t]
\centering
\includegraphics[width=0.96\linewidth]{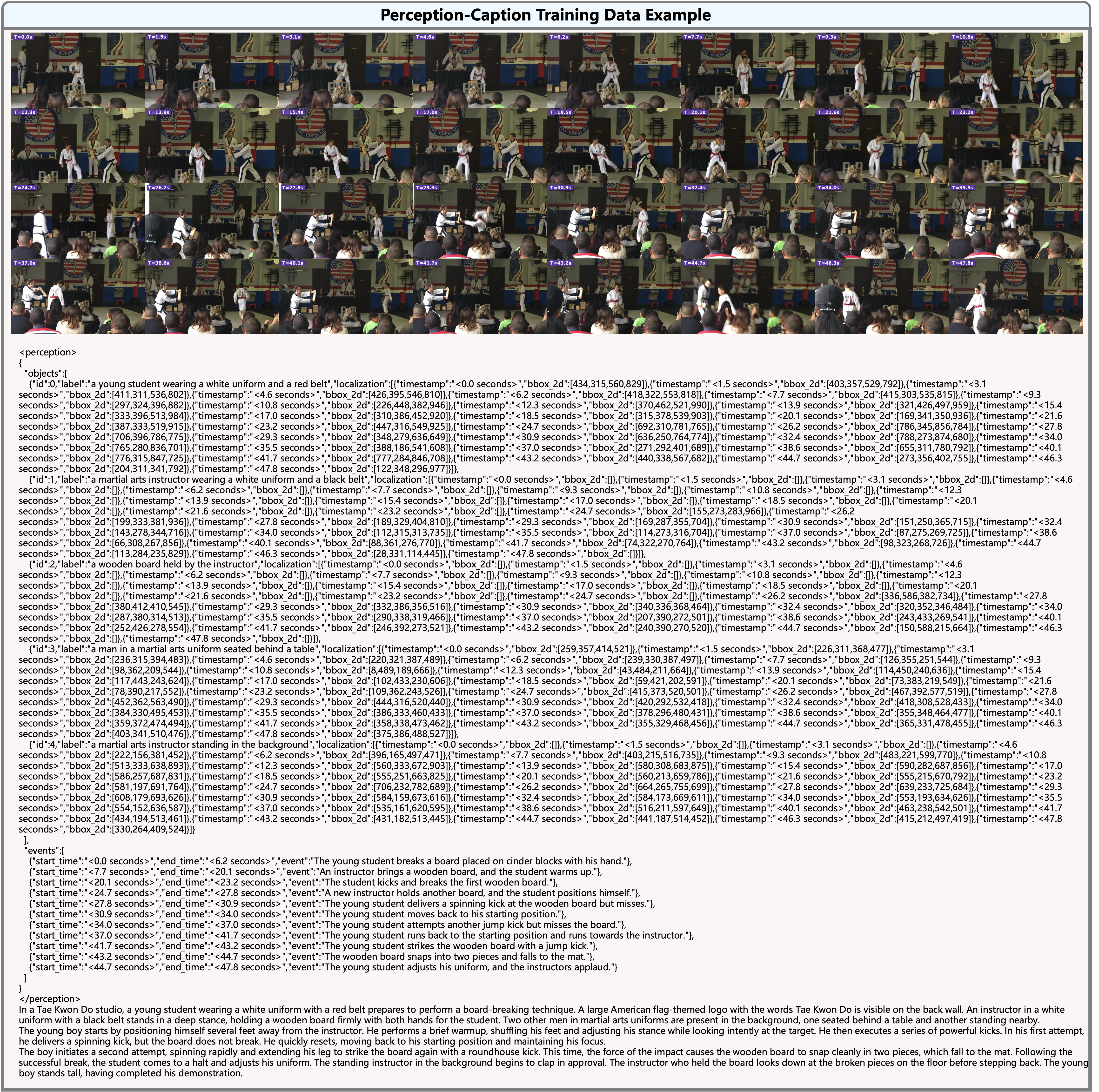}
\caption{Constructed training example from Caption-Anchored Perception Data Construction.}
\label{fig:data_example_2}
\end{figure}

\paragraph{Caption-anchored construction.}
For each video, the annotator model follows four steps:
\begin{enumerate}
    \item \textbf{Caption Anchor Generation.} The annotator model first receives the video in a caption-only setting and generates the final caption $Y^\star_{\mathrm{cap}}$.
    \item \textbf{Object-Event Content Extraction.} The annotator model extracts object labels and event descriptions from $Y^\star_{\mathrm{cap}}$. These extracted textual contents define which entities and events must appear in the perception trace.
    \item \textbf{Video-Grounded Perception Completion.} Conditioned on the video and extracted contents, the annotator model completes the missing perception fields by assigning timestamped boxes to object labels and start/end times to event descriptions.
    \item \textbf{Structured Data Assembly.} The completed objects and events are assembled with the original caption into a structured data instance $Y^\star=[Y^\star_{\mathrm{perc}},Y^\star_{\mathrm{cap}}]$ for PD-SFT and as reference evidence for PG-RL.
\end{enumerate}
This procedure enforces object-event consistency by construction, because the perception trace is grounded from the same caption content that will be used as the final description. Figures~\ref{fig:data_example_1} and \ref{fig:data_example_2} show two constructed training examples, each containing the caption-anchored perception trace and its aligned final caption. All the videos are uniformly sampled with 32 frames during data construction.

\clearpage

\section{Prompt Details}
\label{app:prompt_details}

This section summarizes the prompts used for data construction and inference. The data-construction prompts instantiate the caption-anchored pipeline in Section~\ref{app:data_construction}, while the inference prompts define the caption-only baseline and the perception-aware PercepCap generation format.

\begin{figure}[t]
\centering
\includegraphics[width=0.96\linewidth]{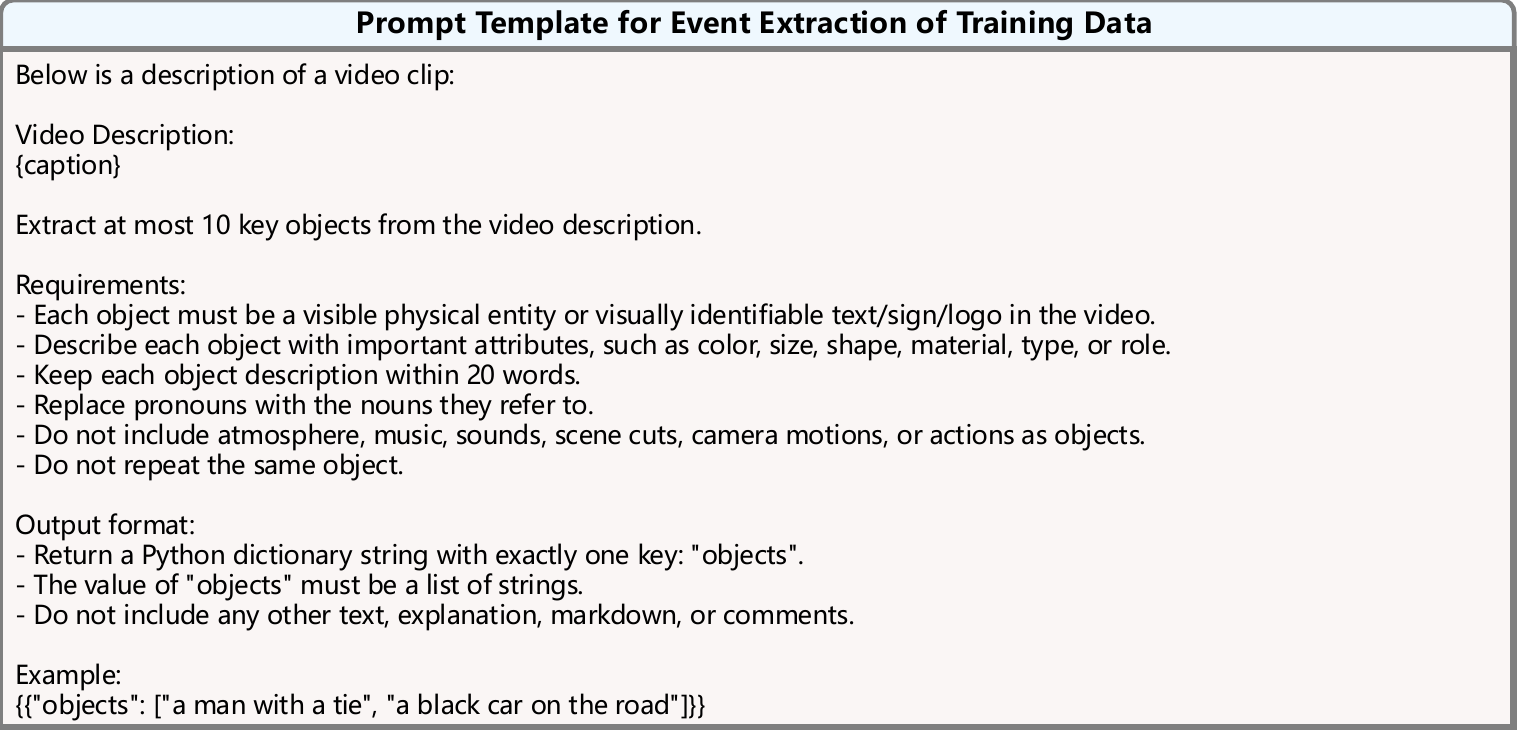}
\caption{Prompt for extracting object labels from the caption during data construction.}
\label{fig:prompt_training_objs}
\end{figure}

\paragraph{Object extraction prompt.}
Figure~\ref{fig:prompt_training_objs} shows the prompt used after caption anchor generation to extract object labels from the caption. The goal is to obtain the entity set that will later be grounded with timestamped boxes, rather than to ask the annotator model to invent additional objects from the video.

\begin{figure}[t]
\centering
\includegraphics[width=0.96\linewidth]{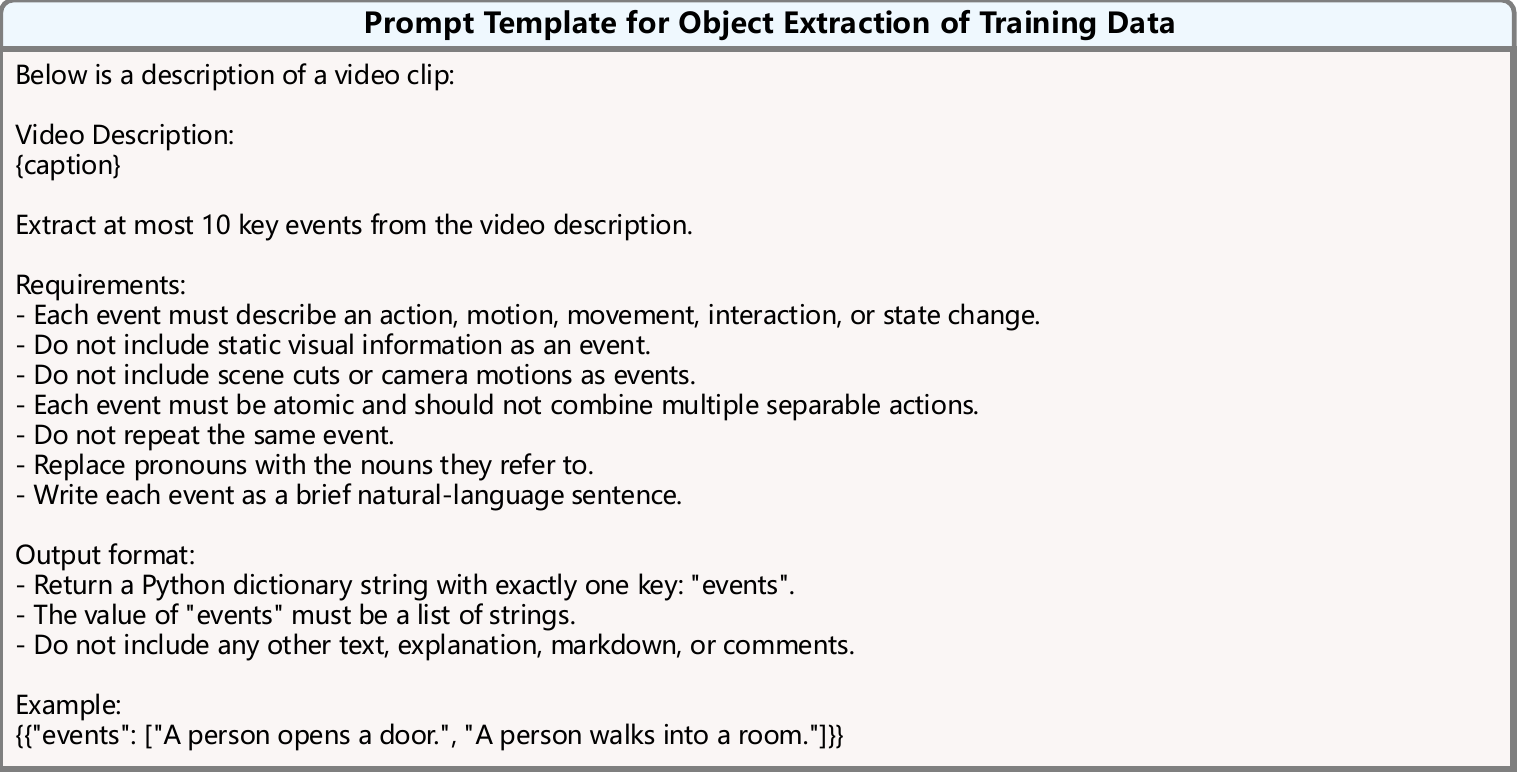}
\caption{Prompt for extracting event descriptions from the caption during data construction.}
\label{fig:prompt_training_events}
\end{figure}

\paragraph{Event extraction prompt.}
Figure~\ref{fig:prompt_training_events} shows the prompt used to extract event descriptions from the caption. These event descriptions define the semantic content whose temporal boundaries are completed in the next construction step.

\begin{figure}[t]
\centering
\includegraphics[width=0.96\linewidth]{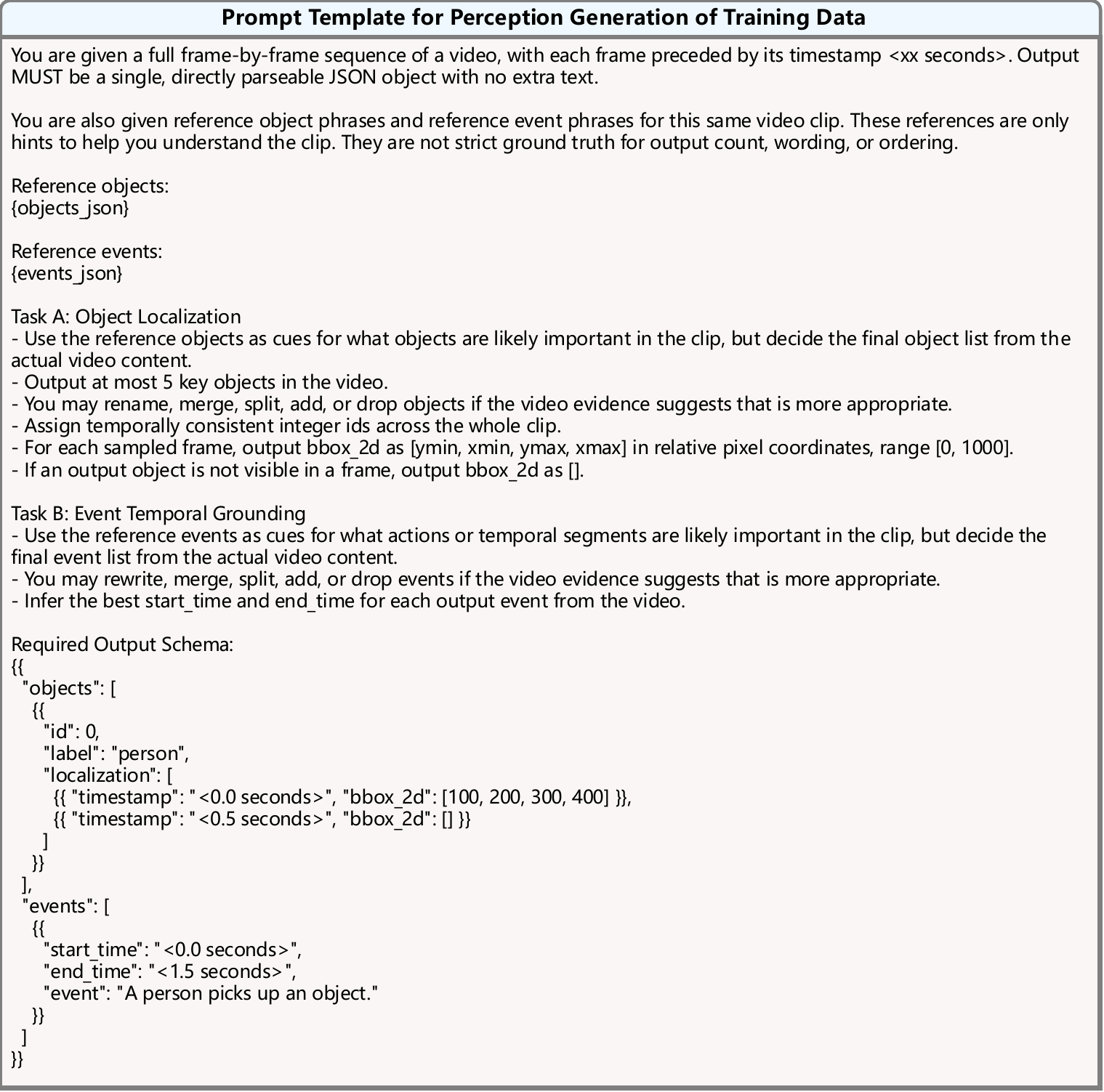}
\caption{Prompt for completing video-grounded perception fields during data construction.}
\label{fig:prompt_training_perception}
\end{figure}

\paragraph{Perception completion prompt.}
Figure~\ref{fig:prompt_training_perception} shows the prompt used to complete the video-grounded perception fields. Given the video together with extracted object labels and event descriptions, the annotator model assigns timestamped boxes to objects and temporal intervals to events.

\begin{figure}[t]
\centering
\includegraphics[width=0.96\linewidth]{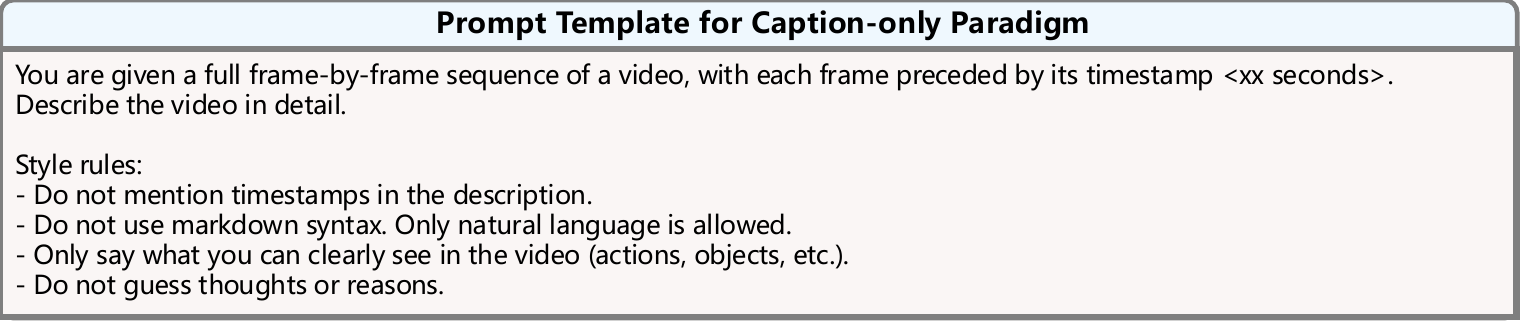}
\caption{Prompt for caption-only inference.}
\label{fig:prompt_caption_only}
\end{figure}

\paragraph{Caption-only inference prompt.}
Figure~\ref{fig:prompt_caption_only} shows the prompt used for the caption-only generation paradigm. It asks the model to produce only the final detailed caption and serves as the baseline inference setting.

\begin{figure}[t]
\centering
\includegraphics[width=0.96\linewidth]{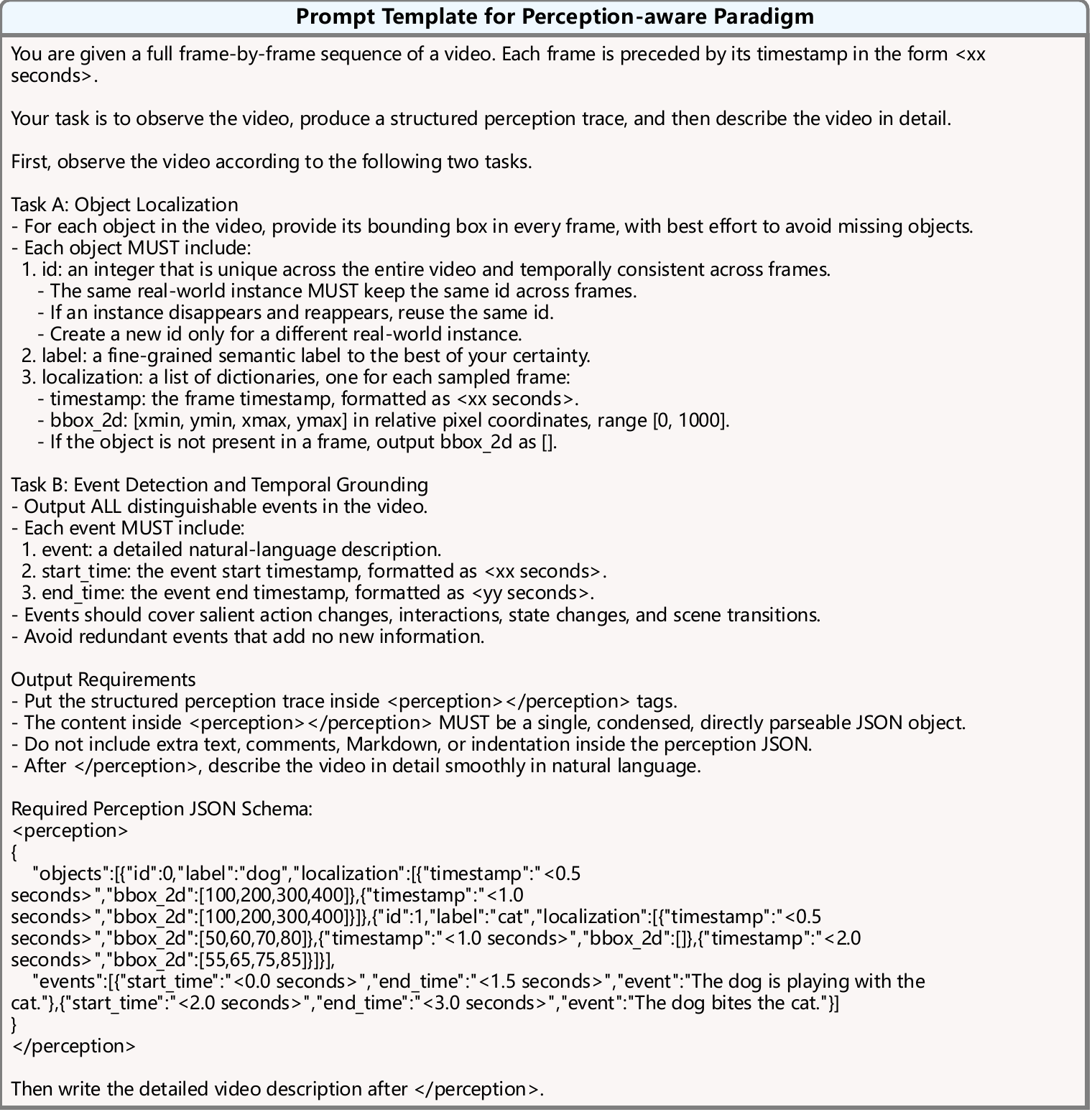}
\caption{Prompt for perception-aware PercepCap inference with structured output.}
\label{fig:prompt_perception_aware_jsonl}
\end{figure}

\paragraph{Perception-aware inference prompt.}
Figure~\ref{fig:prompt_perception_aware_jsonl} shows the prompt used by PercepCap at inference time. It requires the model to produce structured perception before the final caption, enabling the generated trace to be parsed, audited, and used as evidence for caption generation.

\paragraph{Bounding-box coordinate conventions.}
The data-construction annotator and PercepCap intentionally follow their respective models' native bounding-box conventions. The annotator prompt in Figure~\ref{fig:prompt_training_perception} represents boxes as \texttt{[ymin, xmin, ymax, xmax]}, whereas PercepCap training and inference use \texttt{[xmin, ymin, xmax, ymax]}, as shown in Figure~\ref{fig:prompt_perception_aware_jsonl}. Both use relative pixel coordinates in the range $[0,1000]$. Before structured data assembly, we programmatically convert all annotator outputs to \texttt{[xmin, ymin, xmax, ymax]}; PD-SFT supervision, PG-RL reward computation, inference parsing, and evaluation therefore use a single consistent convention.

\clearpage

\section{Broader Impacts}
\label{app:broader_impacts}

PercepCap aims to improve detailed video captioning by making object and event grounding explicit. More grounded captions can benefit video search, accessibility-oriented description, dataset auditing, and analysis tools that require interpretable intermediate evidence rather than only a final text output.

The same capability also carries risks. More accurate video description systems could be misused for privacy-invasive monitoring or surveillance, and incorrect captions may still mislead users if treated as verified facts. The structured perception trace makes some errors easier to inspect, but it does not remove the need for human oversight in high-stakes applications. We therefore view PercepCap as a research system for video understanding and captioning, not as a deployment-ready surveillance or decision-making tool.

\section{Existing Assets and Licenses}
\label{app:asset_licenses}

PercepCap builds on the publicly available Qwen3-VL-8B-Instruct checkpoint~\cite{qwen3technicalreport} and uses Gemini~3 Flash~\cite{gemini3flash} for caption-anchor generation and Gemini~3.1 Pro~\cite{geminiteam2026gemini31problog} for perception-data construction through their provider API and applicable service terms. For evaluation, we use the official benchmarks and procedures of DREAM-1K~\cite{dream1k}, CaReBench~\cite{carebench}, ShortVidBench~\cite{shortvidbench}, MotionBench~\cite{motionbench}, VidCapBench-AE~\cite{vidcapbench}, YouTube-VOS~\cite{youtubevos}, and ActivityNet~\cite{caba2015activitynet}. We credit these assets through their original publications and use them according to their released licenses or terms of use. We do not redistribute third-party benchmark videos, external model checkpoints, or the constructed training data.


\end{document}